\documentclass[10pt,twocolumn,letterpaper]{article}

\usepackage[pagenumbers]{cvpr} % To force page numbers, e.g. for an arXiv version

\usepackage{graphicx}
\usepackage{amsmath}
\usepackage{amssymb}
\usepackage{booktabs}

\usepackage{cite}
\usepackage{times}
\usepackage{epsfig}
\usepackage{graphicx}
\usepackage{amsmath}
\usepackage{amssymb}
\usepackage{acronyms}
\usepackage{xcolor}

\usepackage[accsupp]{axessibility}  % Improves PDF readability for those with disabilities.

\usepackage[font=footnotesize]{caption}
\usepackage[font=footnotesize]{subcaption}

\usepackage{amsmath,amsthm}
\theoremstyle{plain}
\newtheorem{proposition}{Proposition}

\newcommand{\beginsupplement}{
    \renewcommand{\thesection}{\arabic{section}}
    \renewcommand{\theequation}{\arabic{equation}}
    \renewcommand{\thetable}{\arabic{table}}
    \renewcommand{\thefigure}{\arabic{figure}}
}

\usepackage[pagebackref,breaklinks,colorlinks]{hyperref}

\usepackage[capitalize]{cleveref}
\crefname{section}{Sec.}{Secs.}
\Crefname{section}{Section}{Sections}
\Crefname{table}{Table}{Tables}
\crefname{table}{Tab.}{Tabs.}

\def\cvprPaperID{10646} % *** Enter the CVPR Paper ID here
\def\confName{CVPR}
\def\confYear{2023}

\begin{document}

\title{Hyperspherical Embedding for Point Cloud Completion}

% \author{Junming Zhang\\
% University of Michigan\\
% {\tt\small junming@umich.edu}

\author{Junming Zhang, Haomeng Zhang, Ram Vasudevan\\
University of Michigan\\
{\tt\small \{junming,haomeng,ramv\}@umich.edu}
\and
Matthew Johnson-Roberson\\
Carnegie Mellon University\\
{\tt\small mkj@andrew.cmu.edu}
}
\maketitle

%%%%%%%%% ABSTRACT
\begin{abstract}
Most real-world 3D measurements from depth sensors are incomplete, and to address this issue the point cloud completion task aims to predict the complete shapes of objects from partial observations.
Previous works often adapt an encoder-decoder architecture, where the encoder is trained to extract embeddings that are used as inputs to generate predictions from the decoder.
However, the learned embeddings have sparse distribution in the feature space, which leads to worse generalization results during testing.
To address these problems, this paper proposes a hyperspherical module, which transforms and normalizes embeddings from the encoder to be on a unit hypersphere.
With the proposed module, the magnitude and direction of the output hyperspherical embedding are decoupled and only the directional information is optimized.
We theoretically analyze the hyperspherical embedding and show that it enables more stable training with a wider range of learning rates and more compact embedding distributions. 
Experiment results show consistent improvement of point cloud completion in both single-task and multi-task learning, which demonstrates the effectiveness of the proposed method.
The code is available at \url{https://github.com/haomengz/HyperPC}.
\end{abstract}

%%%%%%%%% BODY TEXT

\section{Introduction}
The continual improvement of 3D sensors has made point clouds much more accessible, which drives the development of algorithms to analyze them.
Thanks to deep learning techniques, state of the art algorithms for point cloud analysis have achieved incredible performance~\cite{qi2017pointnet,qi2017pointnet++,shi2019pointrcnn,qi2019deep,hu2021learning} by effectively learning representations from large 3D datasets~\cite{geiger2012we,dai2017scannet,sun2020scalability} and have many applications in robotics, autonomous driving, and 3D modeling.
However, point clouds in the real-world are often incomplete and sparse due to many reasons, such as occlusions, low resolution, and the limited view of 3D sensors.
So it is critical to have an algorithm that is capable of predicting complete shapes of objects from partial observations.

% For example, the predicted complete shapes will provide finer collision-check boundaries than bounding boxes to help autonomous vehicles navigate in narrow lanes or crowded urban regions; compared to incomplete measurements, the robot arm can grasp objects in more reasonable poses with the information of complete shapes, in particular, for those unknown objects. 

\begin{figure}[t!]
    \centering
    \includegraphics[width=0.40\textwidth,height=0.25\textwidth]{{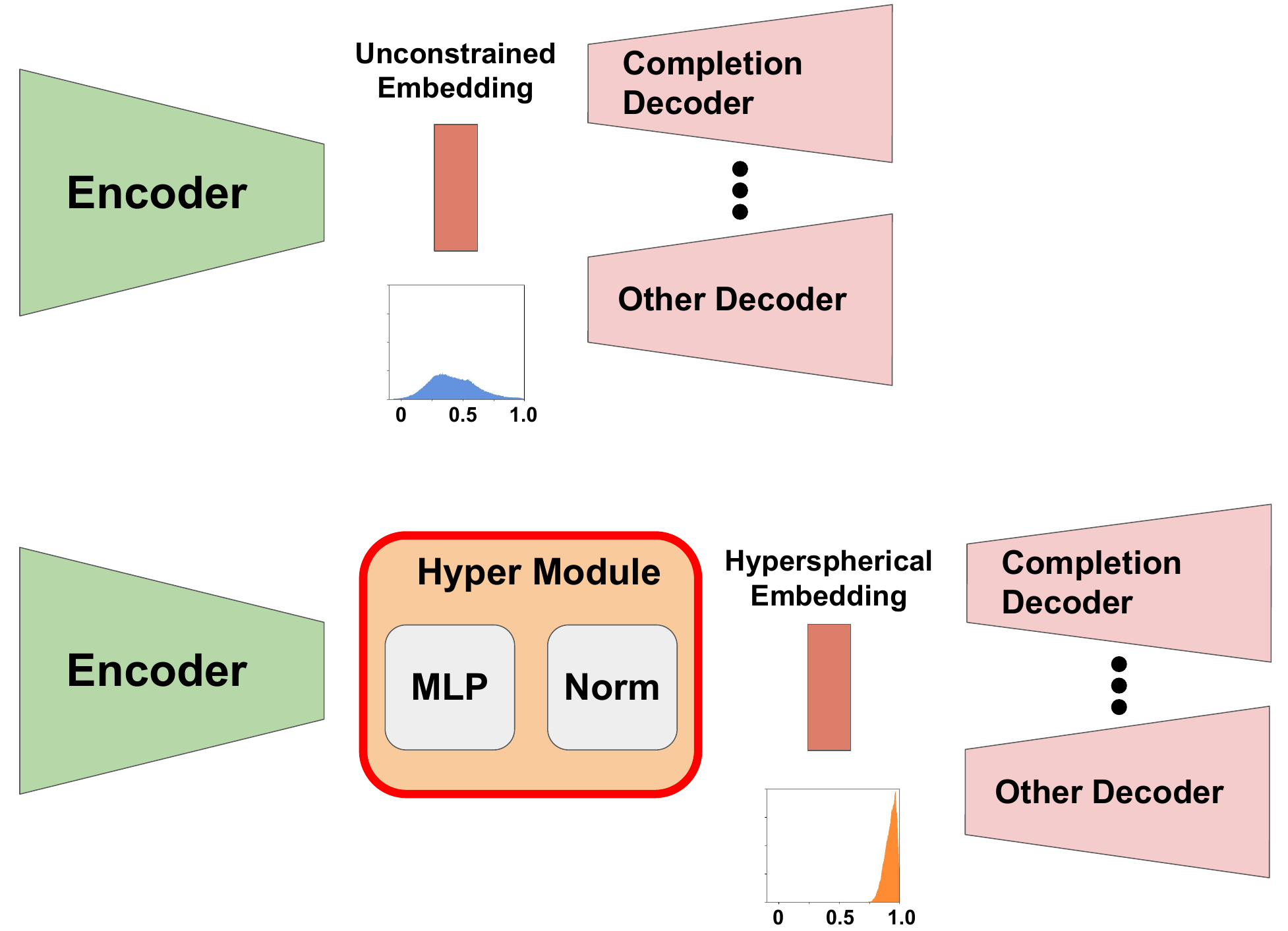}}
    \caption{An illustration of the architecture proposed in this paper. The upper subfigure shows the general point cloud analysis structure, where the embedding is directly output from the encoder without constraints. The lower subfigure shows the structure of the model with the proposed hyperspherical module. The figures under the embeddings illustrate the cosine similarity distribution between embeddings, which indicates a more compact embedding distribution achieved by the proposed method and improves point cloud completion.}
    \label{fig:architecture}
    \vspace{-8mm}
\end{figure}

Given the importance of point cloud completion, it is unsurprising that various methods have been proposed to address this challenge~\cite{yuan2018pcn,tchapmi2019topnet,xiang2021snowflakenet,pan2021variational,wen2020point,wang2020cascaded}.
Most existing methods adapt encoder-decoder structures, in which the encoder takes a partial point cloud as input and outputs an embedding vector, and then it is taken by the decoder which predicts a complete point cloud.
The embedding space is designed to be high-dimensional as it must have large enough capacity to contain all information needed for downstream tasks.
However, the learned high-dimensional embeddings, as shown in this paper, tend to have a sparse distribution in the embedding space, which increases the possibility that unseen features at testing are not captured by the representation learned at training and leads to worse generalizability of models. 

Usually, one real-world application requires predictions from multiple different tasks. 
For example, to grasp an object in space the robot arm would need the information about the shape, category, and orientation of the target object. 
In contrast to training all tasks individually from scratch, a more numerically efficient approach would be to train all relevant tasks jointly by sharing parts of networks between different tasks~\cite{uhrig2016pixel,kendall2018multi,liu2019end}.
However, existing point cloud completion methods lack the analysis of accomplishing point cloud completion jointly with other tasks.
We show that training existing point cloud completion methods with other semantic tasks together leads to worse performance when compared to learning each individually.

To address the above limitations, this paper proposes a hyperspherical module which outputs hyperspherical embeddings for point cloud completion. 
The proposed hyperspherical module can be integrated into existing approaches with encoder-decoder structures as shown in Figure \ref{fig:architecture}.
Specifically, the hyperspherical module transforms and constrains the output embedding onto the surface of a hypersphere by normalizing the embedding's magnitude to unit, so only the directional information is kept for later use.
We theoretically investigate the effects of hyperspherical embeddings and show that it improves the point cloud completion models by more stable training with large learning rate and more generalizability by learning more compact embedding distributions.
We also demonstrate the proposed hyperspherical embedding in multi-task learning, where it helps reconcile the learning conflicts between point cloud completion and other semantic tasks at training.
The reported improvements of the existing state-of-the-art approaches on several public datasets illustrate the effectiveness of the proposed method.
The main contributions of this paper are summarized as follows:
\begin{itemize}
\setlength\itemsep{0em}
\item We propose a hyperspherical module that outputs hyperspherical embeddings, which improves the performance of point cloud completion.

\item We theoretically investigate the effects of hyperspherical embeddings and demonstrate that the point cloud completion benefits from them by stable training and learning a compact embedding distribution. 

\item We analyze training point cloud completion with other tasks and observe conflicts between them, which can be reconciled by the hyperspherical embedding. 
\end{itemize}

\section{Related Work}

\begin{figure*}[t!]
    \centering
    \includegraphics[width=1.0\textwidth]{{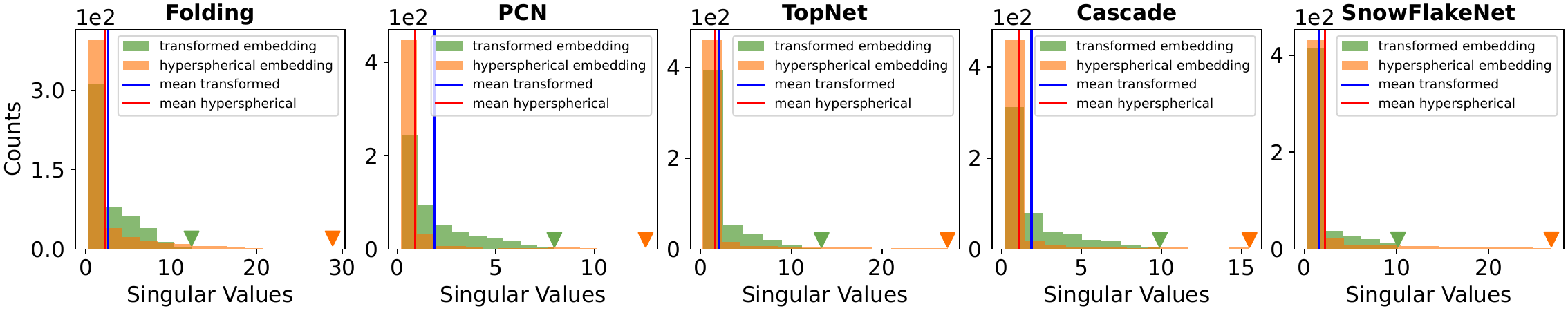}}
    \caption{An illustration of distributions of singular values. We compute singular values of weights in the layer right before the embedding, obtained from the point cloud completion models on MVP dataset with different architectures described in the plot title. The mean of singular values are denoted by vertical lines, and the inverted triangle denotes the largest singular value. This figure shows that the hyperspherical embedding leads to learning more poorly conditioned weights, while the means of singular values with different embeddings are similar.} 
    \label{fig:weight_svd_histogram}
    \vspace{-3mm}
\end{figure*}

\textbf{Point Cloud Completion.}
Point clouds in most 3D datasets ~\cite{geiger2012we,sun2020scalability,dai2017scannet,Chang_2019_CVPR,huang2019apolloscape} are incomplete and sparse due to reasons, such as occlusions, low resolution, and the limited view of 3D sensors.
To address this problem, many works have been proposed to perform point cloud completion, in which people seek to recover the full shape of objects based on the partial observations~\cite{yuan2018pcn,tchapmi2019topnet,zhang2021point, xiang2021snowflakenet,wang2020cascaded,wen2020point}.
Most of them adapt a pipeline of encoding the partial observations into an embedding feature and then decode it to complete point clouds.
Since the structures of encoders have been successfully explored by other 3D tasks~\cite{qi2017pointnet++,qi2017pointnet,wang2019dynamic,liu2019relation}, most completion approaches focus on designing different completion decoders~\cite{yuan2018pcn,yang2018foldingnet,tchapmi2019topnet}.
However, those one-stage decoders are shown to predict point clouds unevenly distributed over the surface of objects and fail to preserve detailed structures in the inputs.
Later approaches address these issues by using a coarse-to-fine decoding strategy, in which the decoding process generates several complete shapes at different resolutions, and the partial point clouds are used along with other intermediate decoding results to maximally preserve structures in the inputs~\cite{xiang2021snowflakenet,wang2020cascaded,wen2020point,wen2021pmp}.
Instead of designing a structure of model, the method developed in the paper focuses on effectively learning representation, which is a more universal problem since the developed method can be integrated into existing structures.
We show the improvement of existing point cloud completion by applying the proposed method on both one-stage and coarse-to-fine decoders. 

\textbf{Embedding in Point Cloud.}
Most methods for point cloud analysis use a max pooling layer as the last layer to address the permutation issue contained in the unordered point set~\cite{qi2017pointnet,qi2017pointnet++,wang2019dynamic,liu2019relation}.
Embeddings output from those encoders are learned from large datasets using an end-to-end training. 
However, the embeddings learned in this way are not imposed with constrain and tend to have a sparse distribution, which increases the possibility of testing inputs accidentally falling into the unseen regions during training due to gaps or holes in the embedding space~\cite{wang2017normface,liu2017sphereface}.
The expressiveness of the embeddings can be improved by other techniques, such as adversarial training~\cite{cai2022learning,wen2021learning}, probabilistic modeling~\cite{pan2021variational,zhang2021point}. 
In multi-task learning, point cloud embeddings are shared by different decoders to improve efficiency of models~\cite{qi2019deep,shi2019pointrcnn,fang2020graspnet}.
Unfortunately, those decoders learned by different task losses may require different embedding distributions, which may lead to optimization conflicts during training, and an unconstrained embedding space makes such issues occur more frequently~\cite{sener2018multi,yu2020gradient}.
In this paper, we propose to use a simple but effective hyperspherical embedding for point cloud completion and demonstrate the success of the proposed method in both single-task learning and multi-task learning.

\textbf{Hyperspherical Embedding.}
Effectively obtaining an embedding that is needed for the corresponding learning task from raw input data is very important.
Compared to the embeddings learned without constraints, some works proposed to use the hyperspherical embedding by normalizing it onto a unit hypersphere and have shown success in many fields, such as representation learning, metric learning, and face verification~\cite{pu2022alignment,wang2020understanding,zhang2020deep,zhang2018heated,liu2017sphereface,wang2017normface,liu2017deep,liu2021learning}.
All those works suggested unit hypersphere is a nice feature space, and some of them tried to explain the improvement by showing the empirical observation of stabler training and better distribution of embeddings when applying hyerspherical embeddings~\cite{zhang2018heated,wang2020understanding,liu2021learning}.
But all of them lack a comprehensive analysis of why the hyperspherical embedding leads to a better distribution of embeddings and the effects in multi-task learning.
Thus, this paper delves into the hyperspherical embedding and we apply it to point cloud completion.
We uncover that the hyperspherical embedding drives the model to learn complexity reduction of high-dimentional data 
 by poorly conditioned weights and thus leads to a compact embedding distribution.

\section{Hyperspherical Embedding for Learning Point Cloud}
\label{sec:method_hyper_embedding}
This section describes the proposed hyperspherical module and investigates effects of the hyperspherical embedding.

\subsection{Proposed Hyperspherical Module}
To address the sparse embedding distribution, we propose a new hyperspherical module and its structure is shown in Figure \ref{fig:architecture}. 
The proposed module contains two layers, a \ac{MLP} layer and a normalization layer.
The outputs from the encoder are first transformed by the \ac{MLP} layer and then the normalization layer constrains the features onto the surface of hypersphere by $l_2$ normalization,
\begin{equation}
    \hat{f} = \frac{f}{\| f \|_2}
\label{eq:normalized_embedding}
\end{equation}
where the $\|f\|_2 = \sqrt{\sum_i f^2_i}$, and $\hat{f}$ denotes the $l_2$ normalized embedding of $f$.

\subsection{Effects of Hyperspherical Embedding}
In this part, we investigate the effects of optimizing the $l_2$ normalized embedding and we draw several conclusions: 
1) the gradient of the embedding before normalization is orthogonal to itself;
2) the magnitude of the embedding before normalization increases at each update during training;
3) the increased magnitude enables more stable training with a wider range of learning rates compared to unconstrained embeddings;
4) the embedding distribution is compact in the angular space, resulting better point cloud completion performance in both single-task and multi-task learning.

\begin{figure}[t!]
    \centering
    \includegraphics[width=0.5\textwidth]{{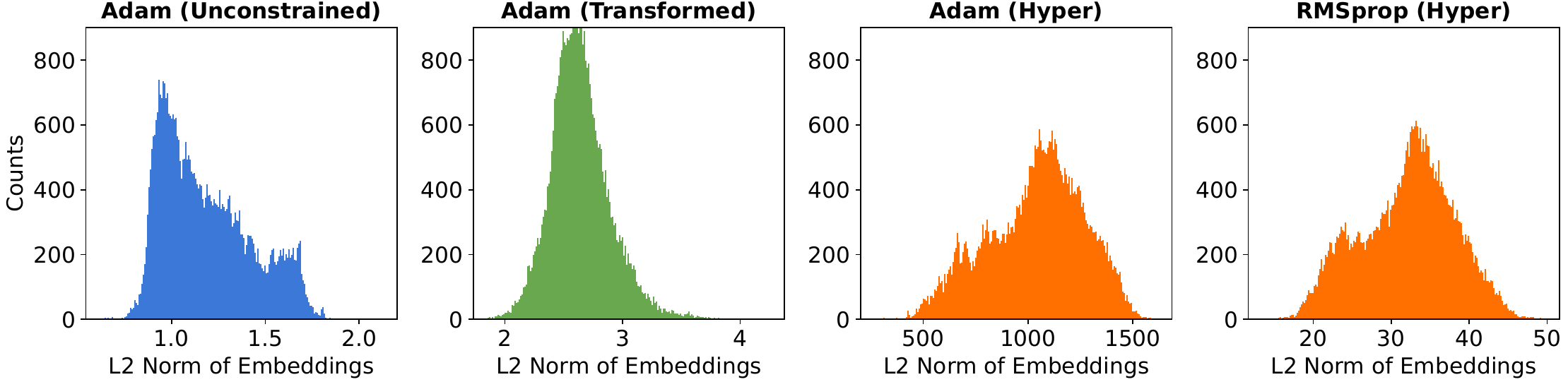}}
    \caption{Norm distribution of embeddings. The embeddings are derived from the MVP test set on point cloud  completion, obtained by using different embeddings or different optimizers, as described in the plot titles. Unconstrained embeddings (Unconstrained) are from models without the proposed module. Transformed embeddings (Transformed) are from models with the proposed module but removing the normalization layer. Hyperspherical embeddings (Hyper) are from models using the proposed module, and the norm shown in this figure is computed before the normalization.}
    \label{fig:embedding_norm_distribution}
\end{figure}

\begin{proposition}
During optimizing $l_2$ normalized embedding $\hat{f}$, the computed gradient to the embedding before normalization denoted by $f$ is orthogonal to itself, $\langle f, \frac{\partial L}{\partial f} \rangle = 0$.
\label{prop:embedding_orthogonal}
\end{proposition}
Proof. 
Suppose the normalization process follows Equation \ref{eq:normalized_embedding} and the loss at optimization is denoted by $L$.
The gradient to embedding $f$ can be formulated as:
\begin{equation}
    \frac{\partial L}{\partial f} = \frac{\frac{\partial L}{\partial \hat{f}} - \hat{f} \langle \frac{\partial L}{\partial \hat{f}}, \hat{f} \rangle}{\| f \|_2}
\label{eq:embedding_gradient}
\end{equation}
Based on it, we can show the orthogonality by computing the inner product between the embedding and its gradient.
More details can be found in the supplementary, and similar conclusions are also reported in~\cite{zhang2020deep,wang2017normface,zhang2018heated}.

% \begin{figure}[t!]
%     \centering
%     \fbox{\includegraphics[width=0.45\textwidth, height=0.2\textwidth]{{figures/empty.pdf}}}
%     \caption{Illustration of the normalization operation and its gradient in 2-dimensional space}
%     \label{fig:illustration_normalization_and_gradient}
% \end{figure}

\begin{proposition}
For standard stochastic gradient descent (SGD), each update of embedding $f$ will monotonically increase its norm, $\| f \|_2$.
\label{prop:embedding_norm_increase}
\end{proposition}
Proof. The orthogonality between the embedding and its gradient from Proposition \ref{prop:embedding_orthogonal} indicates that applying gradient at each update increases the norm of an embedding, which is validated by showing the distribution of embedding’s norm in Figure \ref{fig:embedding_norm_distribution}.
Similar observations are reported in~\cite{zhang2020deep,zhang2018heated}.
This property is based on the vanilla gradient descent algorithm and does not strictly hold for optimizers that use momentum or separate learning rates for individual parameters.
However, we still find the same effect empirically hold for other SGD-based optimizers, such as Adam~\cite{kingma2014adam} and RMSprop~\cite{hinton2012neural}, as illustrated in Figure \ref{fig:embedding_norm_distribution}.

\begin{proposition}
The magnitude of the gradient is inversely proportional to the norm of the embedding, $ \frac{\partial L}{\partial f} \propto \frac{ 1 }{\| f \|_2}$.
\label{prop:gradient_norm_embedding}
\end{proposition}
Proof. Considering the norm $\|f\|_2$ in the denominator in Equation \ref{eq:embedding_gradient}, the magnitude of the gradient is inversely proportional to the norm of an embedding.
This conclusion is similar to the one in~\cite{zhang2018heated}.
However, we further note that this effect enables optimizing neural networks with a wider range of learning rates.
In particular, the norm of embedding trained with a large learning rate quickly increases shown by Proposition \ref{prop:embedding_norm_increase} until an appropriate effective learning rate is reached, while the same setting puts the model at risk of overshooting the minima.
So it may not be able to converge when using unconstrained embeddings.
Normalizing the weights of neural networks has similar effects reported by~\cite{salimans2016weight}, while in this paper we focus on normalizing the layer of embedding and keep weights untouched, which makes the implementation easier.
Figure \ref{fig:mvp_multitask} shows the comparison results with hyperspherical embedding and unconstrained embeddings trained using different learning rates, and more results can be found in Figure \ref{fig:modelnet40_multitask} and Figure \ref{fig:shapenet_multitask} in the supplementary.
All of them show that the hyperspherical embedding leads to more stable training with a wider range of learning rates and better performance of point cloud completion.

\begin{proposition}
Considering a vector is transformed by a matrix, $i.e.$, $f = Wx$. During optimization, the increased norm of $f$ requires a poorly conditioned matrix $W$.
\label{prop:embedding_weights_singular}
\end{proposition}
Proof. The matrix $W$ can be decomposed by \ac{SVD}:
\begin{equation}
    W = U\Sigma V^{T}
    \label{eq:svd}
\end{equation}
where $U$ and $V$ are orthonormal matrices and $\Sigma$ is a diagonal matrix.
Based on Equation \ref{eq:svd}, the norm of the transformed output $f$ is only related to singular values contained in $\Sigma$, since orthonormal matrices $U$ and $V$ do not modify the magnitude of inputs.
In our case, the norm of $f$ increases during training from Proposition \ref{prop:embedding_norm_increase}, so the $\|\Sigma\|_2$ will increase during optimization.
However, increasing all singular values in $\|\Sigma\|$ makes the weight large, which adversely affects the performance of models by overfitting the training data~\cite{goodfellow2016deep}.
Empirically we do not observe this issue and find that the mean of the singular values trained with normalized embeddings stays similarly to the one without using it, shown in Figure \ref{fig:weight_svd_histogram}.
Therefore, the increase of certain singular values in $\Sigma$ will inevitably lead to decrease of other singular values. 
In particular, the large singular values get increased and small singular values get decreased to increase the $\|\Sigma\|_2$, which causes the weight $W$ to be poorly conditioned as it is illustrated in Figure \ref{fig:weight_svd_histogram}.

% due to the fixed sum of them by the Frobenius norm of $W$.
% It is known that Frobenius norm of $W$ is equal to the $l_2$ norm of singular values
% \begin{equation}
%     \|W\|_2 = \sum_{i=1}^{n} \sum_{j=1}^{m} W_{ij} = \| \Sigma \|_2
% \end{equation}
% so the increase of certain singular values in $\Sigma$ will inevitably lead to decrease of other singular values, which causes the weight $W$ to be more singular as it is illustrated in Figure \ref{fig:weight_svd_histogram}.
% Actually, the assumption in Proposition \ref{prop:embedding_weights_singular} does not strictly hold in practice since no such constraint is applied at training. 
% However, the weights in neural networks are prevented from being large due to the regularization techniques~\cite{}, which makes the weights in the model smaller and helps neural networks reduce overfitting issues.
% Empirically, we found that the norm of weights tend to change stably and slowly at training as shown by Figure \ref{fig:weight_norm_at_training} , which helps the assumption loosely hold. 

% \begin{figure}[t!]
%   \centering
%   \includegraphics[width=0.45\textwidth, height=0.3\textwidth]{{figures/weight_norm_at_training.pdf}}
%   \caption{The average $L2$ norm of weights in the model at training.}
%   \label{fig:weight_norm_at_training}
% \end{figure}

\begin{figure}[t!]
    \centering
    \includegraphics[width=0.45\textwidth]{{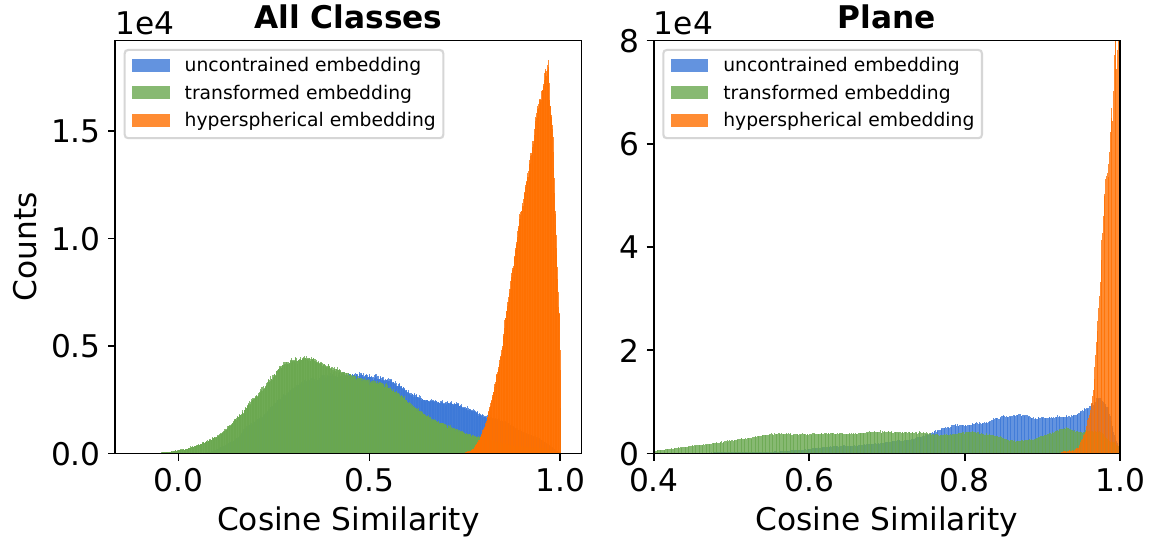}}
    \caption{Cosine similarity distribution of embeddings. We compute pairwise cosine distance between embeddings obtained from the test set in MVP dataset. We visualize the distribution in either one class or overall classes as described in the plot titles. It shows that hyperspherical embeddings have more compact angular distribution. More visualizations of different classes can be found in Figure \ref{fig:cosine_angular_distribution_all_class} in the supplementary.}
    \label{fig:embedding_angular_distribution}
    \vspace{-10mm}
\end{figure}

One effect of poorly conditioned weights is that it will reduce the complexity in the input high-dimensional data by keeping information on principle axes while ignoring information on other axes. 
By doing this, transformed vectors by those weights tend to point in a similar direction.
Moreover, the hyperspherical embedding follows a $l_2$ normalization, so the discrepancy of embeddings in magnitude is further removed.
We compute pairwise cosine distance of embeddings trained with or without $l_2$ normalization in the test set and visualize the distribution in Figure \ref{fig:embedding_angular_distribution}.
It shows that hyperspherical embeddings have a more compact angular distribution, while the unconstrained embedding distribution tends to be sparse.
This compact embedding distribution helps the model generalize well on unseen data at testing and increases the generalizability of models.

\begin{figure}[t]
    \centering
    \includegraphics[width=0.5\textwidth]{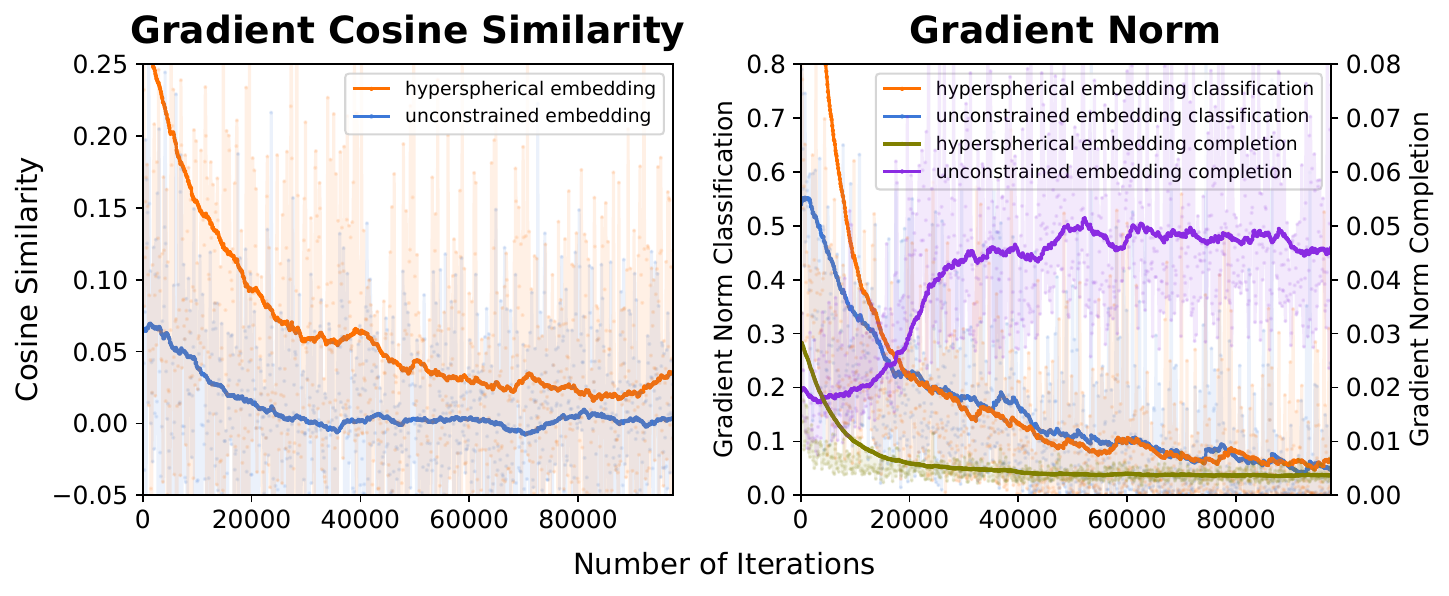}
    \caption{An illustration of gradient conflicts between tasks in multi-task learning during training. We visualize the gradient cosine similarity and gradient magnitude as indicated by the titles of subfigures, obtained by training point cloud completion and shape classification on MVP dataset. Hyperspherical embeddings lead to smaller gradient conflicts between tasks in multi-task learning.}
    \label{fig:cosine_similarity_gradients_between_tasks}
    \vspace{-5mm}
\end{figure}

The learned compact embedding distribution also helps reconcile the learning conflicts in multi-task learning.
The resulting compact embedding distribution forces different tasks to learn within the shared space, while the unconstrained embedding space provides tasks the freedom to land on optimal embedding distributions with discrepancy. 
We use the gradient cosine similarity proposed by~\cite{yu2020gradient} to measure the conflicts between different tasks and visualize the training process in Figure \ref{fig:cosine_similarity_gradients_between_tasks}. 
The figure shows that the hyperspherical embeddings lead to smaller gradient conflicts at training, as larger cosine similarity indicates smaller gradient conflicts.
More discussion about the effects on multi-task learning can be found in Sec \ref{sec:experiments_effect_hyper_embedding}.

\section{Experiments}
Experiments are divided into three parts.
We first report results on different datasets to evaluate the effectiveness of the proposed methods in Sec \ref{sec:experiments_different_datasets}.
Second, we conduct a detailed ablation study to validate the design of our hyperspherical module in Sec \ref{sec:ablation_study}.
Finally, we analyze and visualize the effects of hyperspherical embeddings introduced in Sec \ref{sec:experiments_effect_hyper_embedding}. 
All experiments are conducted on NVIDIA Tesla V100 GPUs, and we use default training settings for all baseline methods. 
In multi-task learning experiments, we use fully connected layers and cross entropy loss to train semantic tasks in multi-task learning, while decoders with more sophisticated structures are used to generate complete point clouds trained using Chamfer Distance~\cite{fan2017point}.

\begin{figure*}[t!]
    \centering
    \includegraphics[width=0.85\textwidth, height=0.32\textwidth]{{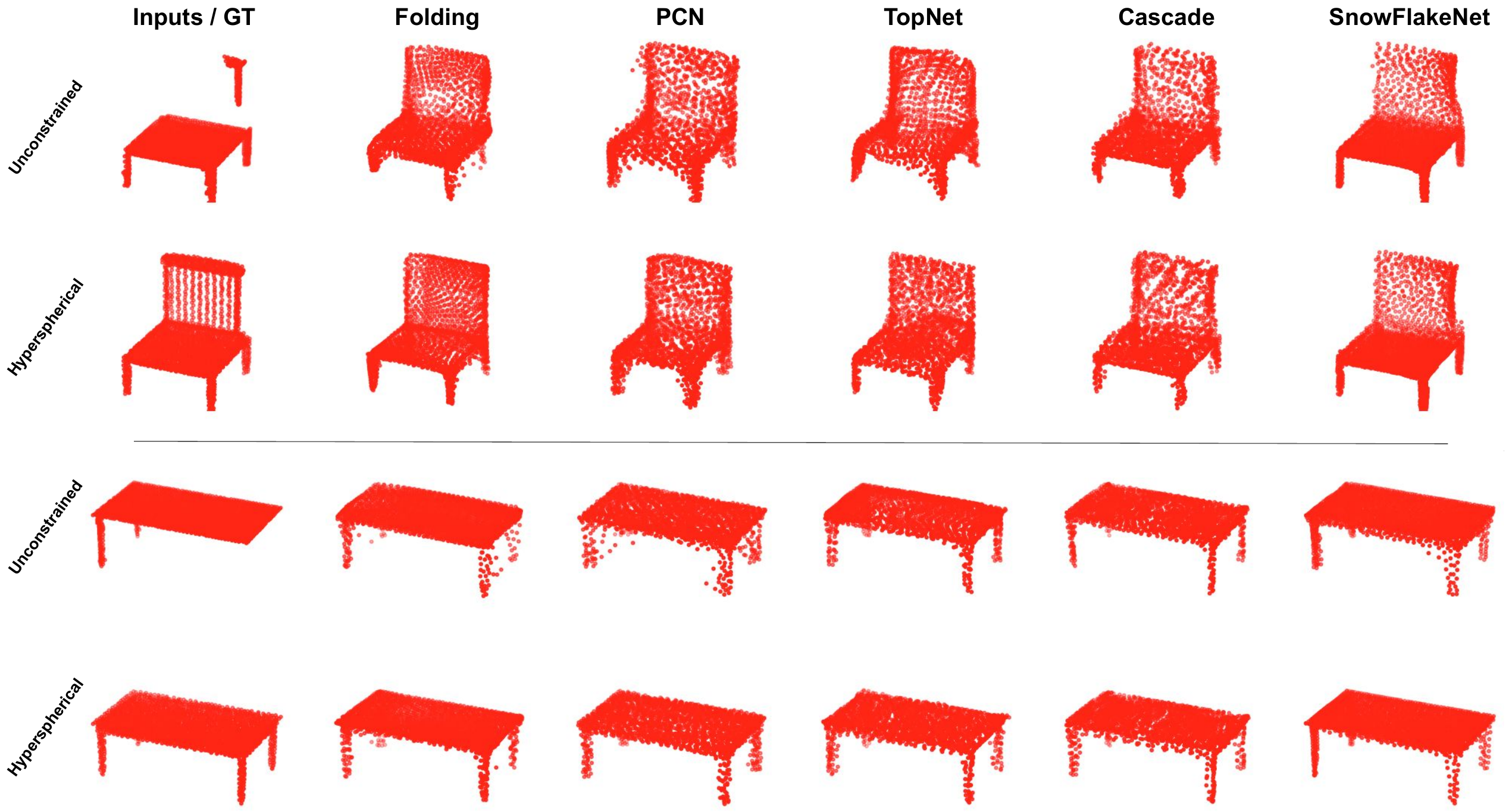}}
    \caption{Qualitative results of various state-of-the-arts point cloud completion approaches on MVP test set.}
    \label{fig:mvp}
\end{figure*}

\subsection{Experiments on Different Datasets}
\label{sec:experiments_different_datasets}

\begin{table*}[t!]
\centering
\scalebox{0.7}{
    \begin{tabular}{l|cccccccccccccccc|c}
    Model & plane & cabinet & car & chair & lamp & sofa & table & wcraft & bed & bench & bshelf & bus & guitar & mbike & pistol & sboard & average \\
    \hline\hline
    Folding~\cite{yang2018foldingnet} & 4.71 & 9.08 & 6.81 & 15.22 & 23.12 & 10.28 & 14.32 & 9.90 & 22.02 & 10.28 & 14.48 & 5.24 & \textbf{2.02} & 6.91 & 7.21 & 4.59 & 10.39 \\
    Folding (H) & \textbf{4.48 }& \textbf{8.82 }& \textbf{6.68 }& \textbf{13.79} & \textbf{21.44} & \textbf{9.66 }& \textbf{12.98} & \textbf{8.57 }& \textbf{18.93} & \textbf{8.96 }& \textbf{13.44} & \textbf{4.95 }& 2.03 & \textbf{6.43 }& \textbf{6.22 }& \textbf{4.20 }& \textbf{9.47 }\\
    \hline
    PCN~\cite{yuan2018pcn} & \textbf{4.23} & 9.35 & 6.73 & 13.56 & \textbf{20.94} & 10.51 & 14.20 & 9.81 & 21.32 & 9.98 & 15.08 & 5.45 & 1.90 & 6.23 & 6.23 & 5.03 & 10.03 \\
    PCN (H) & 4.24 & \textbf{9.14 }& \textbf{6.49 }& \textbf{13.04} & 22.47 & \textbf{10.04} & \textbf{12.99} & \textbf{8.75 }& \textbf{18.95} & \textbf{9.33 }& \textbf{13.93} & \textbf{5.06 }& \textbf{1.84 }& \textbf{6.00 }& \textbf{5.92 }& \textbf{4.15 }& \textbf{9.52 }\\
    \hline
    TopNet~\cite{tchapmi2019topnet}  &  4.63 & 9.23 & 6.79 & 14.31 & 19.50 & 10.48 & 14.30 & 9.65 & 20.54 & 10.12 & 15.53 & 5.36 & 2.09 & 6.77 & 7.74 & 4.94 & 10.12 \\ 
    TopNet (H) & \textbf{4.07 }& \textbf{9.13 }& \textbf{6.75 }& \textbf{13.08} & \textbf{19.45} & \textbf{10.03} & \textbf{12.85} & \textbf{8.89 }& \textbf{19.50} & \textbf{9.63 }& \textbf{14.33} & \textbf{5.23 }& \textbf{2.03 }& \textbf{6.66 }& \textbf{6.42 }& \textbf{3.92 }& \textbf{9.50 }\\
    \hline
    Cascade~\cite{wang2020cascaded} & 2.66 & 8.69 & 6.02 & 10.22 & 13.07 & 8.76 & 9.90 & 6.67 & 16.44 & 7.56 & \textbf{11.00} & 4.97 & 1.98 & 4.58 & 4.54 & 2.78 & 7.49 \\
    Cascade (H) & \textbf{2.61 }& \textbf{8.52 }& \textbf{5.97 }& \textbf{9.52 }& \textbf{12.03} & \textbf{8.71 }& \textbf{9.83 }& \textbf{6.46 }& \textbf{15.78} & \textbf{7.17 }& 11.15 & \textbf{4.90 }& \textbf{1.88 }& \textbf{4.50 }& \textbf{4.24 }& \textbf{2.76 }& \textbf{7.25 }\\
    \hline
    SnowFlakeNet~\cite{xiang2021snowflakenet} & 1.94 & 7.61 & 5.61 & 6.77 & \textbf{6.82} & 7.09 & 7.21 & \textbf{4.65} & 10.98 & 4.76 & 7.54 & 4.16 & 1.14 & 3.78 & 3.15 & \textbf{2.67} & 5.37 \\
    SnowFlakeNet (H) & \textbf{1.89 }& \textbf{7.26 }& \textbf{5.36 }& \textbf{6.50 }& 7.59 & \textbf{6.72 }& \textbf{6.63 }& 4.67 & \textbf{10.39} & \textbf{4.39 }& \textbf{7.37 }& \textbf{4.03 }& \textbf{0.95 }& \textbf{3.60 }& \textbf{3.15 }& 2.84 & \textbf{5.21 }\\

    \end{tabular}
    }\caption{The performance of different completion approaches trained on MVP dataset. 
    The Chamfer Distance is reported, multiplied by $10^4$, on the provided test set.
    ``H" indicates using the proposed hyperspherical module.}
    \label{table:completion_mvp}
    \vspace{-3mm}
\end{table*}

\begin{figure}[t!]
    \centering
    \includegraphics[width=0.48\textwidth]{{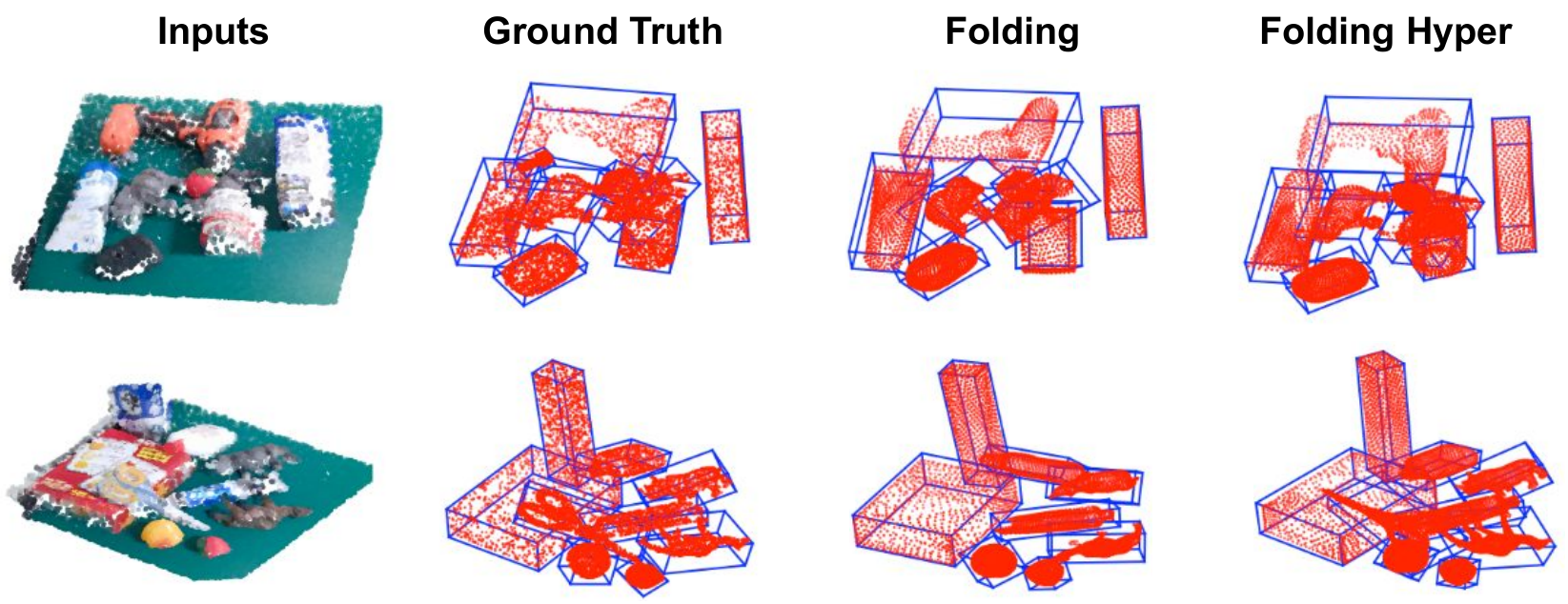}}
    \caption{Qualitative 3D detection, pose estimation, and point cloud completion results on GraspNet test set.}
    \label{fig:graspnet}
\end{figure}

Note, a longer description of the considered datasets and experimental results on \textbf{ModelNet40} and \textbf{ShapeNet} can be found in the supplementary document. 

\textbf{MVP}\; 
We evaluate point cloud completion on MVP~\cite{pan2021variational} and compare the results of generating complete shape with 2048 points, and qualitative results can be found in Figure \ref{fig:mvp}. 
Compared to the predictions without our hyperspherical module, the predicted point clouds tend to be a bit more blurry and contain more noise points.
We assume it is due to the test embeddings falling into regions not close to features captured at training.
More quantitative results are shown in Table \ref{table:completion_mvp}.
We compare the baseline models by adding the hyperspherical module after the encoder and denote them by (H).
One exception is SnowFlakeNet, in which the decoding process also consists of encoding modules, so we add hyperspherical modules in its decoder as well.
To achieve fair comparison, we train all methods with the best training setting and report their results.
Table \ref{table:completion_mvp} shows that using our proposed module brings consistent improvements of all existing completion approaches by 3 $\sim$ 9\% decrease of average class Chamfer Distance.
Multi-task learning on MVP dataset will be discussed in Sec \ref{sec:experiments_effect_hyper_embedding}.

\textbf{GraspNet}\;
To demonstrate on real-world scenarios, in this part we aim to detect objects in 3D space along with predicting their complete shapes on GraspNet dataset~\cite{fang2020graspnet}.  
We modified the structures of VoteNet~\cite{qi2019deep}, which was designed to detect 3D objects from point clouds. 
The input point clouds are converted from the depth image captured by RGBD sensors.
After extracting the embedding from each proposal, three branches of decoders are followed to generate predictions of 6DoF 3D bounding boxes, semantics and objectness scores, and complete point clouds of objects, respectively.
In the evaluation, object detection is measured by mean average precision (mAP), poses of objects are measured by the symmetry metric proposed in~\cite{xiang2017posecnn}, and the point cloud completion is measured by Chamfer Distance.
We evaluate the effectiveness of the proposed hyperspherical module in this multi-task learning scenario, and the results are shown in Table \ref{table:graspnet}.
Since the model structures are the same except for decoders, we distinguish the models by the decoders in the first column.
Two-stage completion decoders, such as Cascade and SnowFlakeNet, are demonstrated to have better performance on synthetic dataset, but they refine on perfect partial point clouds located on object's surfaces, which are inaccessible in this case. 
As it is shown in Table \ref{table:graspnet}, hyperspherical embeddings help all three metrics with noticeable improvement comparing to unconstrained embeddings. 
Qualitative results can be found in Figure \ref{fig:graspnet} and Figure \ref{fig:graspnet_more} in the supplementary.

\begin{figure*}[t!]
    \centering
    \includegraphics[width=0.95\textwidth,height=0.2\textwidth]{{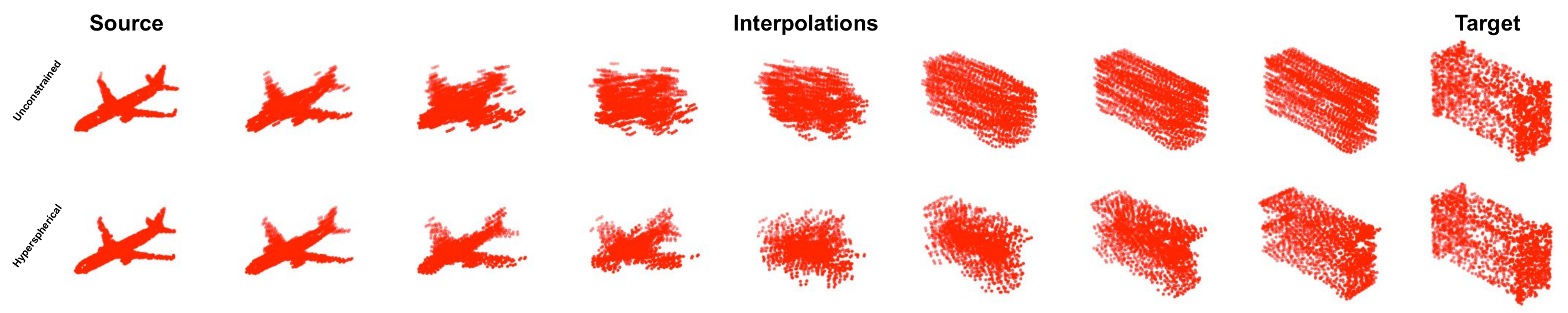}}
    \caption{Illustration of point cloud interpolation in the embedding space. The generated point clouds with hyperspherical embeddings have more clear clues from source or target shapes than those with unconstrained embeddings.}
    \label{fig:interpolation}
    \vspace{-5mm}
\end{figure*}

\begin{table}[t]
\centering
\scalebox{1.0}{
    \begin{tabular}{l|ccc}
    Model & mAP (0.25) & CD & Pose Acc.\\
    \hline\hline
    Folding &  70.50 & 0.18 & 52.42 \\ 
    Folding (H) &  \textbf{71.21} & \textbf{0.14} & \textbf{54.01} \\ 
    \hline
    PCN & 69.11 & 0.21 & 50.33 \\ 
    PCN (H) & \textbf{70.93} & \textbf{0.15} & \textbf{52.39} \\ 
    \end{tabular}
    }\caption{Performance on GraspNet test set.
    Average precision with 3D IoU threshold 0.25 (mAP 0.25) is reported for object detection, Chamfer Distance (CD) is reported for point cloud completion, multiplied by $10^4$, and pose accuracy (Pose Acc.) is reported for 6D pose estimation.
    The first column indicates the structure of decoder used in the model, and ``H" indicates using the proposed hyperspherical module.}
    \label{table:graspnet}
\end{table}

\begin{table}[t!]
\centering
\scalebox{0.7}{
    \begin{tabular}{ccccccc|c}
    Base & 1-layer MLP & 2-layer MLP & ReLU\&BN & $l_1$ & $l_2$ & $l_3$ & CD\\
    \hline\hline
    \checkmark & & & & & & & 10.39 \\
    \checkmark & \checkmark & & & & & & 10.12 \\
    \checkmark & & & & & \checkmark & & 9.57 \\    
    \checkmark & \checkmark & & & & \checkmark & & \textbf{9.47} \\
    \checkmark & \checkmark & & \checkmark & & \checkmark & & 10.06 \\     
    \checkmark & & \checkmark & & & \checkmark & & 10.02 \\
    \checkmark & \checkmark & & & & & \checkmark & 9.69 \\
    \checkmark & \checkmark & & & \checkmark & & & 9.99 \\
     \hline
    \end{tabular}
    }\caption{Results of ablation study. The reported metric is Chamfer Distance, multiplied by $10^4$, of point cloud completion on MVP test set. ``Base" indicates not using the proposed module, or unconstrained embeddings; ``1-layer MLP" indicates using one MLP layer in the hyperspherical moduel; ``2-layer MLP" indicates using two MLP layers in the hyperspherical moduel; ``ReLU\&BN" indicates using ReLU activation and Batch Normalization after each MLP layer; ``L*" indicates the type of normalization in the hyperspherical module.}
    \label{table:ablation_study}
\end{table}

% \begin{table}[t!]
% \centering
% \scalebox{0.65}{
%     \begin{tabular}{cccccc|c}
%     Unconstrained & Transformed & Hyperspherical & Reg & BN & 2-layer MLP & CD\\
%     \hline\hline
%     \checkmark & & & & & & 10.39 \\
%      & \checkmark & & & & & 10.12 \\
%      & & L2 & & & & \textbf{9.47} \\
%      & & L1 & & & & 9.99 \\
%      & & L3 & & & & 9.69 \\
%      \hline
%      & & L2 & 0.001 & & & 10.4 \\
%      & & L2 & 0.01 & & & 10.73 \\
%      & & L2 & 0.1 & & & 10.89 \\
%      \hline \hline
%      & \checkmark & & & & \checkmark & 10.86 \\
%      & & L2 & & & \checkmark & \textit{\textbf{10.32}} \\
%      \hline
%      & \checkmark & & & \checkmark & & 10.57 \\
%      & & L2 & & \checkmark & & \textit{\textbf{10.06}} \\
%      \hline
%      & \checkmark & & & \checkmark & \checkmark & 11.86 \\
%      & & L2 & & \checkmark & \checkmark & \textit{\textbf{10.57}} \\
%     \end{tabular}
%     }\caption{Results of ablation study. The reported metric is Chamfer Distance, multiplied by $10^4$, of point cloud completion on MVP test set. The first three column indicates the results of models with different types of embeddings described similarly in Figure \ref{fig:embedding_norm_distribution}. ``reg" indicates training with regularization term to discourage large norms of embeddings, and the number next to it indicates the weights when adding the term to training loss.}
%     \label{table:ablation_study}
%     \vspace{-3mm}
% \end{table}

\subsection{Ablation Study}
\label{sec:ablation_study}
Results of the ablation study are shown in Table \ref{table:ablation_study} and numbers in the table are the average class Chamfer Distance of completion models with folding decoder reported on the MVP test set.
The accuracy of the model is improved after applying the proposed hyperspherical module, reducing the Chamfer Distance from $10.39$ to $9.47$, shown by row one and row four.
The structural design of the hyper module is validated by results in the first four rows, and they also indicates that the major improvement is brought by the normalization layer, from $10.12$ to $9.47$.
From row three and row five, changing MLP layers, activation layer or batch normalization layer does not improve the performance of point cloud completion compared to the design used in this paper.
To validate the choice of $l_2$ manifold, we also report the results of applying $l_1$ ($9.99$) and $l_3$ ($9.69$) normalization.
All of them are shown to boost the performance, but the $l_2$ manifold used in this paper has the best results.

\subsection{Experiments on the Effects of Hyperspherical Embedding}
\label{sec:experiments_effect_hyper_embedding}

In this section, we study the effects of hyperspherical embedding and provide empirical results and visualizations that align with the conclusions claimed in Sec \ref{sec:method_hyper_embedding}.

\textbf{Increased Magnitude of Embedding}\;
From Proposition \ref{prop:embedding_norm_increase}, the magnitude of an embedding gets increased during the optimization, and we visualize the distribution of embedding magnitude from test set on MVP dataset in Figure \ref{fig:embedding_norm_distribution}.
When optimizing our proposed hyperspherical embedding denoted by (Hyper), the scale of embedding's magnitude before normalization is significantly larger than those unnormalized denoted by (Unconstrained and Transformed).
Furthermore, we observe in two leftmost subfigures that the range of embeddings' magnitude changes little after adding a \ac{MLP} layer, which validates that $l2$ normalization is the key to increase embedding magnitude.
Even though the effect is based on vanilla gradient descent, we still find it empirically holds for modern optimizers, such as Adam and RMSProp shown in the two rightmost subfigures of Figure \ref{fig:embedding_norm_distribution}.

\begin{figure*}[t!]
    \centering
    \includegraphics[width=1.0\textwidth]{{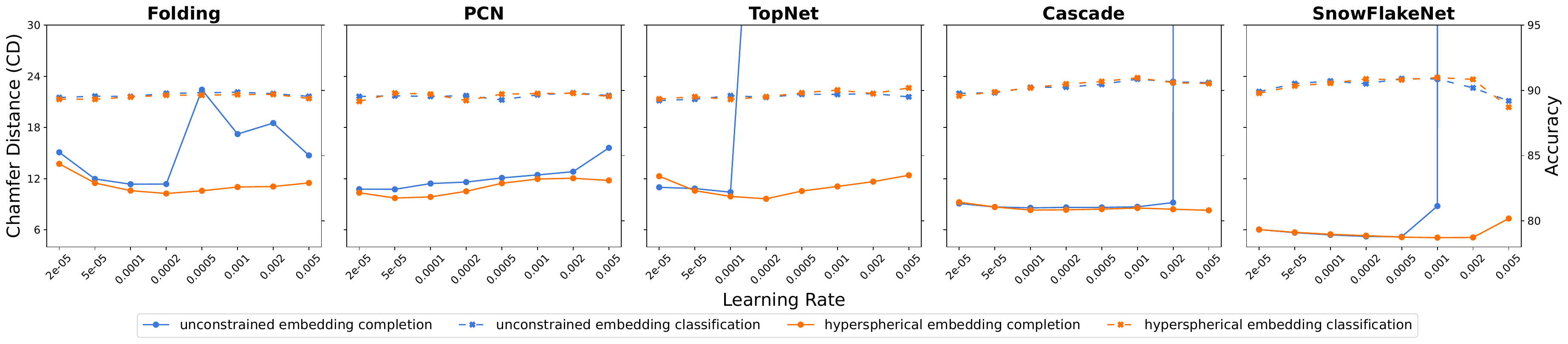}}
    \caption{Performance of multi-task learning of point cloud completion and classification on MVP dataset with different learning rates.}
    \label{fig:mvp_multitask}
    \vspace{-3mm}
\end{figure*}

\textbf{Enabling Wider Range of Learning Rates}\;
When optimizing a normalized embedding, the gradient at each update can be computed by following the Equation \ref{eq:embedding_gradient}, and it indicates that the magnitude of gradient is inversely proportional to the norm of the embedding shown in Proposition \ref{prop:gradient_norm_embedding}.
One benefit of this finding is that the increased norms of embeddings help neural networks gain robustness to varying values of learning rate.
Figure \ref{fig:mvp_multitask} (Figure \ref{fig:modelnet40_multitask} and Figure \ref{fig:shapenet_multitask} are in the supplementary) show the results of models on different tasks using different learning rates.
Compared to the point cloud completion results with unconstrained embedding, the models using hyperspherical embeddings perform more stably under a wider range of learning rates with better performance in particular for large learning rates.
If the learning rate is large, the norm of embeddings before normalization quickly increases and makes the effective learning rate small enough to stabilize training.
When trained with small learning rates, models perform similarly using either hyperspherical embeddings or unconstrained embeddings.
This can be explained by the observation that slightly increased or similar norms of embeddings using the hyperspherical embeddings compared to unconstrained embeddings, and a small learning rate would slow down the speed of increase of embeddings' norm.

\begin{table*}[t!]
\centering
\scalebox{0.8}{
    \begin{tabular}{l|cc||cc|cc|cc|cc|c}
     & \multicolumn{2}{|c||}{Single Task} & \multicolumn{2}{|c|}{Equal Weights} & \multicolumn{2}{|c|}{PCGrad~\cite{yu2020gradient}} & \multicolumn{2}{|c|}{Uncert. ~\cite{kendall2018multi}} & \multicolumn{2}{|c|}{Weight Search}\\
    Model & Acc & CD & Acc & CD & Acc & CD & Acc & CD & Acc & CD & S. vs. M.\\
    \hline\hline
    Folding & 89.68 & 10.39 & 89.77 & 11.37 & 89.67 & 11.21 & 89.81 & 11.22 & 89.12 & 10.45 & -0.58\\ 
    Folding (H) & 89.91 & \textbf{9.47} & 89.63 & \textbf{10.26} & 89.77 & \textbf{10.13} & 89.51 & \textbf{10.07} & 89.43 & \textbf{9.40} & \textbf{0.74} \\
    \hline
    PCN & 89.62 & 10.03 & 89.58 & 10.75 & 89.41 & 10.75 & 89.33 & 10.77 & 89.26 & 10.37 & -3.39\\ 
    PCN (H) & 89.55 & \textbf{9.52} & 89.79 & \textbf{9.73} & 89.56 & \textbf{9.58} & 89.69 & \textbf{9.58} & 89.78 & \textbf{9.45} & \textbf{0.74}\\ 
    \hline
    TopNet & 89.49 & 10.12 & 89.59 & 10.42 & 89.84 & 10.33 & 89.58 & 10.52 & 89.43 & 10.24 & -1.19\\ 
    TopNet (H) & 89.55 & \textbf{9.50} & 89.51 & \textbf{9.64} & 89.80 & \textbf{9.59} & 89.90 & \textbf{9.48} & 89.74 & \textbf{8.79} & \textbf{7.47}\\ 
    \hline
    Cascade & 90.91 & 7.49 & 90.23 & 8.58 & 90.33 & 8.53 & 90.27  & 8.51 & 90.18  & 7.50 & -0.13\\ 
    Cascade (H) & 90.51 & \textbf{7.25} & 90.19 & \textbf{8.32} & 90.02 & \textbf{8.18} & 90.32 & \textbf{8.32} & 90.48 & \textbf{7.22}& \textbf{0.41} \\ 
    \hline
    SnowFlakeNet & 90.93 & 5.37 & 90.90 & 5.19 & 90.99 & 5.29 & 90.18  & 5.27 & 90.75  & 5.04 & \textbf{6.15}\\ 
    SnowFlakeNet (H) & 90.91 & \textbf{5.21} & 90.98 & \textbf{5.09} & 90.95 & \textbf{5.21} & 90.13 & \textbf{5.11} & 90.82 & \textbf{5.02} & 3.65\\ 
    \end{tabular}
    }\caption{Comparison results of models using different multi-task training strategies on MVP dataset. Results of shape classification (Acc) and point cloud completion (CD) are reported, multipied by $10^4$. ``S. vs. M." shows the percentage of performance change comparing best completion results in multi-task learning to those in single-task learning.}
    \label{table:multitask}
    \vspace{-5mm}
\end{table*}

\textbf{Compact Embedding Distribution}\;
Based on Proposition \ref{prop:embedding_norm_increase} and Proposition \ref{prop:embedding_weights_singular}, optimizing hyperspherical embedding drives the model to learn complexity reduction by ill-conditioned weights.
We empirically verify this by visualizing the singular value distribution of weights trained with point cloud completion task on MVP dataset, and the results are shown in Figure \ref{fig:weight_svd_histogram}.
We observe that singular values of weights learned with the hyperspherical embedding have a larger value span than those without normalized embedding, while the majority of them are located near zero.

The resulting poorly conditioned weights make a compact embedding distribution in the angular space.
Figure \ref{fig:embedding_angular_distribution} illustrates the angular distribution of embeddings by computing cosine similarity of pairwise embeddings from the MVP test set.
Adding a \ac{MLP} layer to transform the unconstrained embedding does not change the span of angular distribution significantly, while the hyperspherical embeddings have much narrower angular span and are distributed more compactly on both single class and overall classes.
Compared to a compact embedding distribution, one disadvantage of the sparse embedding distribution is that it increases the possibility of unseen features falling far away from the seen features at training, which inevitably worsens the generalization of models at testing.
To visually demonstrate the degree of sparsity in the embedding space, we use trained models that perform point cloud reconstruction on ModelNet40 to interpolate embeddings on the embedding space and generate point clouds, and the results are shown in Figure \ref{fig:interpolation}.
Compared to results from unconstrained embedding space, the point clouds in hyperspherical embeddings space have more clear clues from source or target shapes, because the interpolated hyperspherical embeddings are closer to features captured at training.
Thus, it helps with generating more reasonable shapes of objects at testing.

\textbf{Improvement on Multi-task Learning}\;
% In addition to single-task learning, the proposed hyperspherical module also improves the performance of point cloud completion in multi-task learning with other semantic tasks.
% To verify this, we report the results of multi-task learning on three different datasets: ModelNet40, ShapeNet, and MVP.
% We report the results of jointly training point cloud completion and shape classification on MVP in Figure~\ref{fig:mvp_multitask}.
% Since the ModelNet40 and ShapeNet do not provide pairs of partial and complete point clouds, results of point cloud reconstruction are reported along with shape classification on ModelNet in Figure \ref{fig:modelnet40_multitask} (in the supplementary) and part segmentation on ShapeNet in Figure \ref{fig:shapenet_multitask} (in the supplementary).
% By comparing the results of models with unconstrained embeddings, the proposed hyperspherical module has little benefit on the performance of semantic tasks.
% However, the proposed module improves the point cloud completion task with more stable performance when using large learning rates than those with unconstrained embeddings, since the same settings tend to cause the unconstrained models unconverged.
% In terms of the converged results, models with our method still outperform their baselines with noticeable improvement.
In addition to single-task learning, the proposed hyperspherical module also improves the performance of point cloud completion in multi-task learning with other semantic tasks.
To verify this, we report the results of jointly training point cloud completion and shape classification on MVP dataset in Figure~\ref{fig:mvp_multitask}.
More results of multi-task learning on ModelNet40 and ShapeNet datasets can be found in the supplementary. 
By comparing the results of models with unconstrained embeddings, the proposed hyperspherical module has little benefit on the performance of semantic tasks.
However, the proposed module improves the point cloud completion task with more stable performance when using large learning rates than those with unconstrained embeddings, since the same settings tend to cause the unconstrained models unconverged.
In terms of the converged results, models with our method still outperform their baselines.

To make a fair comparison, we also report results of other approaches developed for multi-task learning using different types of embeddings trained on MVP dataset as shown in Table \ref{table:multitask}.
The second to fifth columns show results of models with different training strategies indicated by the column title.
Unsurprisingly, models with the proposed hyperspherical modules outperform the baselines on point cloud completion in both single-task and multi-task learnings under all settings.
By comparing the completion results in multi-task learning, manually searching the optimal weights between completion task and classification task takes much time, but it achieves the best completion results with little affection on classification performance.
Furthermore, the rightmost column (S. vs. M.) presents the percent of changes comparing the best completion results in multi-task learning to those in single-task learning.
It shows improvement of completion performance trained in multi-task learning when using our method, while the models with unconstrained embeddings struggle in the degradation of completion performance when they are trained with shape classification.
One exception is SnowFlakeNet with unconstrained embeddings, which gets improved on completion when trained with classification.
We argue that it is due to multi-task learning helps with reducing the overfitting issue observed when training SnowFlakeNet in single-task learning, but our proposed method (5.02) still outperforms its baseline (5.04). 

This paper focuses on point cloud completion, but the empirical results show a neutral effect on classification.
Table \ref{table:multitask} shows slight improvement (Folding and TopNet) and regression (PCN, Cascade and SnowFlakeNet) in single-task classification.
Hyperspherical embedding leads to a compact embedding space, which means that the inter-class space is reduced compared to regular embedding space.
We suppose that it raises challenges when learning a classifier and explains that hyperspherical embedding is not commonly adopted in single-task learning of classification.
However, our method helps classification to be robust in multitasking while we observe noticeable degradation of classification without our method in weight search column.

To delve into the effects of hyperspherical embedding on multi-task learning, we visualize the conflicts between tasks when training them jointly in Figure \ref{fig:cosine_similarity_gradients_between_tasks}.
Specifically, the measure we visualize is the cosine similarity of gradients on the shared encoders with respect to different task losses, where negative values indicate conflicting gradients, as proposed by~\cite{yu2020gradient}.
The visualization shows that the gradient cosine similarity with hyperspherical embeddings are almost positive, while those with unconstrained embeddings tend to be more negative, which explains the improvement of point cloud completion in multi-task learning with classification observed in Table \ref{table:multitask}.
Moreover, we find that the classification task dominates the training since its magnitude of gradient is significantly larger than gradients with respect to completion loss, as shown in the right subfigure in Figure \ref{fig:cosine_similarity_gradients_between_tasks}.
This aligns well with the results of weight search experiments, where smaller classification weight or larger completion weight lead to better completion performance.

\section{Conclusions}
This paper proposes a general module for point cloud completion.
In particular, our hyperspherical module transforms and normalizes the output from the encoder onto the surface of a hypersphere before it is processed by the following decoders.
We study the effects of the proposed hyperspherical embeddings in both theoretical and experimental ways.
Extensive experiments are performed on synthetic and real-world datasets, and the achieved state-of-the-art results in both single-task and multi-task learnings demonstrate the effectiveness of the proposed method.

\noindent \textbf{Acknowledgements}. This work is supported by the Ford Motor Company via award N022977 and the National Science Foundation under award
1751093.

%%%%%%%%% REFERENCES
{\small
\bibliographystyle{ieee_fullname}
% \bibliography{egbib}
\bibliography{reference}
}

\clearpage
\beginsupplement
\section{Supplementary}
This document provides additional proof, technical details, quantitative results, and qualitative visualization to the main paper.

\subsection{Proof of Proposition}

Suppose the normalization process follows Equation \ref{eq:normalized_embedding}, and the loss at optimization is denoted by $L$, and the gradient to embedding $f$ follows Equation \ref{eq:embedding_gradient}.
Based on them, we show the orthogonality between an embedding and its gradient by computing the their inner product:
\begin{equation}
\begin{split}
    \langle f, \frac{\partial L}{\partial f} \rangle & = \frac{\langle f,\frac{\partial L}{\partial \hat{f}} \rangle - \langle f, \hat{f} \rangle \langle \frac{\partial L}{\partial \hat{f}}, \hat{f} \rangle}{\| f \|_2} \\
    & = \frac{\langle f,\frac{\partial L}{\partial \hat{f}} \rangle - \langle \hat{f}, \hat{f} \rangle \langle \frac{\partial L}{\partial \hat{f}}, f \rangle}{\| f \|_2} \\
    & = \frac{\langle f,\frac{\partial L}{\partial \hat{f}} \rangle - \langle \frac{\partial L}{\partial \hat{f}}, f \rangle}{\| f \|_2} \\
    & = 0
\end{split}
\label{eq:orthogonality_embedding_and_gradient}
\end{equation}

\subsection{Description of Dataset}
\textbf{ModelNet40}\;
ModelNet40 dataset contains 12,311 shapes from 40 object categories, and they are split into 9,843 for training and 2,468 for testing.
Since the dataset does not provide partial point clouds, we evaluate our proposed method on performing point cloud reconstruction and shape classification.
We generate input point clouds by evenly sampling 1024 points from the surface of objects and normalize them within a unit sphere, and no data augmentation is used at training.

\textbf{ShapeNet}\;
ShapeNet part dataset~\cite{wu20153d} dataset contains 16,881 shapes from 16 object categories with 50 parts.
Each point cloud contains 2048 points which are generated by evenly sampling from the surface of objects, and we follow the same set splitting as in~\cite{qi2017pointnet++}. 
Since the dataset does not provide partial point clouds, we evaluate our proposed method on performing point cloud reconstruction and part segmentation.

\textbf{MVP}\; 
MVP dataset~\cite{pan2021variational} contains pairs of partial and complete point clouds from 16 categories.
Partial point clouds are generated by back-projecting 2.5D depth images into 3D space and complete point clouds are used as ground truth.
In the experiments, we apply the set splitting given by the dataset and no data augmentation is used.

\textbf{GraspNet}\;
Most point cloud completion approaches reported results on synthetic datasets, since collecting a real-world dataset with annotation of complete shapes is expensive.
Unfortunately, incomplete measurements from real-world 3D sensors differ from those point clouds synthesized in a simulated environment, and approaches trained on synthetic dataset struggle when tasked to perform completion in real-world scenarios. 
More recently, the GraspNet~\cite{fang2020graspnet} dataset was released and it contains the groundtruth complete shapes of objects, which helps evaluate point cloud completion. 
GraspNet contains 190 cluttered and complex scenes captured by RGBD cameras, bringing 97,280 images in total.
For each image, the accurate 6D pose and the dense grasp poses are annotated for each object.
There are in total 88 objects with provided CAD 3D models, and we use them to generate complete shape groundtruth with 1024 points.

\begin{table}[t]
\centering
\small
\scalebox{1.0}{
    \begin{tabular}{l|cc}
    Model & Seg Acc. & CD \\
    \hline\hline
    PointNet-Folding & 92.06 & 50.08\\ 
    PointNet-Folding (H)& 92.01 & \textbf{34.75} \\ 
    \hline
    PointNet-PCN & 92.06 & 43.61 \\
    PointNet-PCN (H) & 92.01 & \textbf{38.18} \\ 
    \hline
    PointNet-TopNet & 92.06 & 37.4 \\
    PointNet-TopNet (H) & 92.01 & \textbf{35.50} \\
    \hline \hline
    DGCNN-Folding & 92.50 & 49.21\\ 
    DGCNN-Folding (H) & 92.39 & \textbf{33.88} \\ 
    \hline
    DGCNN-PCN & 92.50 & 42.42\\
    DGCNN-PCN (H)& 92.39 & \textbf{37.11} \\ 
    \hline
    DGCNN-TopNet & 92.50 & 36.80 \\
    DGCNN-TopNet (H) & 92.38 & \textbf{35.10}\\
    \end{tabular}
    }\caption{Single-task learning on ShapeNet. Overall point segmentation accuracy (Seg Acc.) is reported for part segmentation, and Chamfer Distance (CD) is reported for point cloud reconstruction, multiplied by $10^4$. The first column describes the encoders and decoders used in the model, and ``H" indicates using the proposed hyperspherical module.}
    \label{table:shapenet}
\end{table}

\begin{table}[t]
\centering
\small
\scalebox{1.0}{
    \begin{tabular}{l|cc}
    Model & Cls Acc. & CD \\
    \hline\hline
    PointNet-Folding & 87.33 & 75.86\\ 
    PointNet-Folding (H) & 87.36 & \textbf{48.88}\\ 
    \hline
    PointNet-PCN & 87.33 & 48.17\\
    PointNet-PCN (H) & 87.36 & \textbf{43.55}\\ 
    \hline
    PointNet-TopNet & 87.33 & 55.04\\
    PointNet-TopNet (H) & 87.36 & \textbf{49.65}\\
    \hline \hline
    DGCNN-Folding & 89.22 & 70.32\\ 
    DGCNN-Folding (H) & 89.47 & \textbf{45.37}\\ 
    \hline
    DGCNN-PCN & 89.22 & 46.54\\
    DGCNN-PCN (H) & 89.47 & \textbf{42.70}\\ 
    \hline
    DGCNN-TopNet & 89.22 & 55.87\\
    DGCNN-TopNet (H) & 89.47 & \textbf{48.75} \\
    \end{tabular}
    }\caption{Single-task learning on ModelNet40.
    Overall classification accuracy (Cls Acc.) is reported for shape classification, and Chamfer distance (CD) is reported for point cloud reconstruction, multiplied by $10^4$.
    The first column describes the encoders and decoders used in the model, and ``H" indicates using the proposed hyperspherical module.}
    \label{table:modelnet40}
\end{table}

\begin{figure*}[h]
    \centering
    \includegraphics[width=0.9\textwidth]{{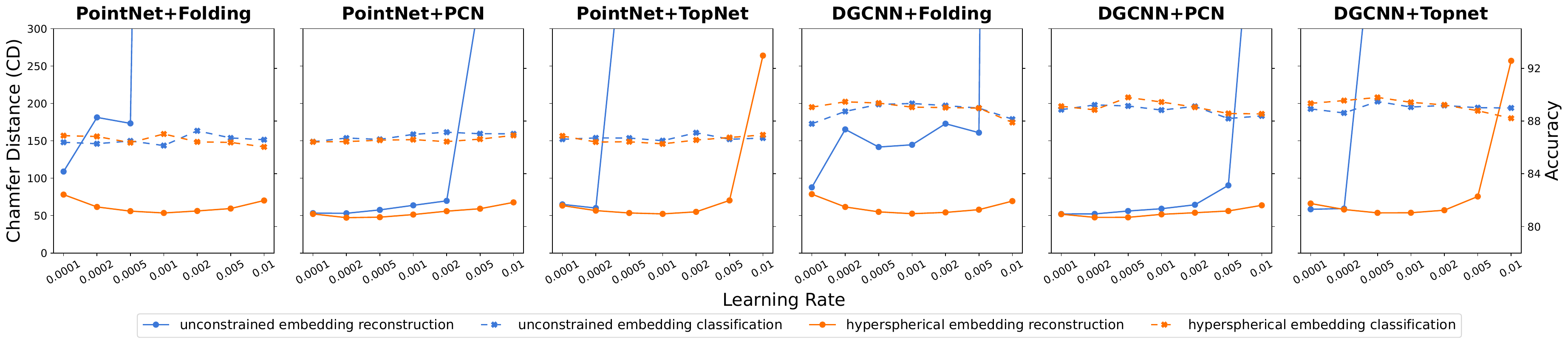}}
    \caption{Performance of multi-task learning of point cloud reconstruction and classification on ModelNet40 with different learning rates.}
    \label{fig:modelnet40_multitask}
\end{figure*}

\subsection{More Experiments}
Since ModelNet40 and ShapeNet do not provide pairs of partial and complete point clouds, we report results of point cloud reconstruction along with shape classification on ModelNet40 and part segmentation on ShapeNet.

\textbf{ModelNet40}\;
Quantitative results on Modelnet40~\cite{wu20153d} are shown in Table \ref{table:modelnet40}.
To make a fair comparison, we report results of the combination of two popular encoders, PointNet~\cite{qi2017pointnet} without T-Net and DGCNN~\cite{wang2019dynamic}, and three different point cloud decoders, Folding~\cite{yang2018foldingnet}, PCN~\cite{yuan2018pcn} and TopNet~\cite{tchapmi2019topnet}. 
The baseline models are compared to their variants added with our proposed hyperspherical modules denoted with (H).
As shown by the last column in Table \ref{table:modelnet40}, our proposed hyperspherical module helps baseline approaches gain noticeable decrease of Chamfer Distance in all cases.
We also test the proposed module in shape classification by removing point cloud decoders and adding fully connected layers.
As shown in the second column, the proposed method module leads to slightly better performance of shape classification. 
Multi-task learning results on shape reconstruction and classification are shown in Figure \ref{fig:modelnet40_multitask}.
By comparing the results of models with unconstrained embeddings, the proposed hyperspherical module have little effect on the performance of semantic tasks.
However, models with the proposed module have more stable performance when using large learning rates than those with unconstrained embeddings, since the same setting tend to cause the training with unconstrained embeddings unconverged.
In terms of converged results, models with our method still outperform their baselines with noticeable improvement.

\begin{figure*}[t!]
    \centering
    \includegraphics[width=0.9\textwidth]{{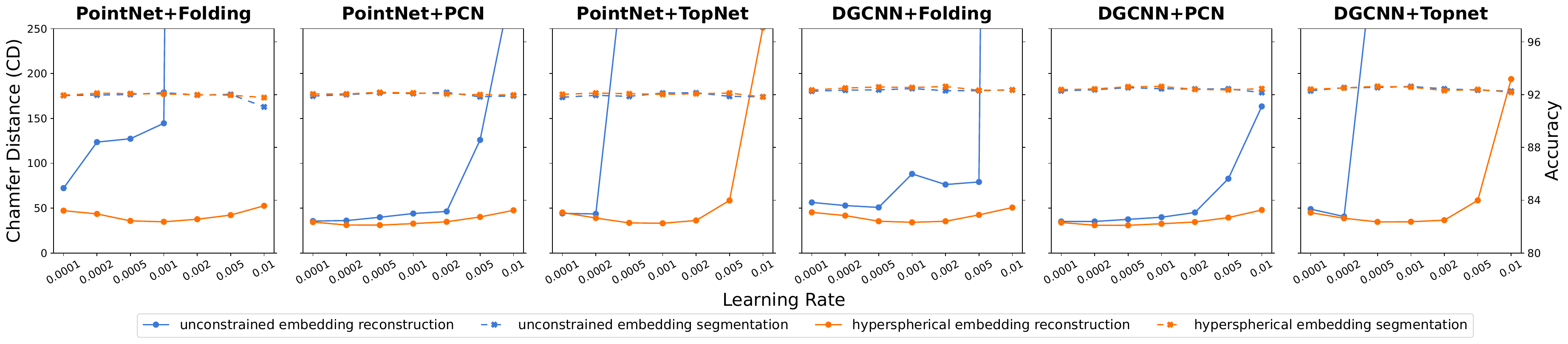}}
    \caption{Performance of multi-task learning of point cloud reconstruction and part segmentation on ShapeNet with different learning rates.}
    \label{fig:shapenet_multitask}
\end{figure*}

\textbf{ShapeNet}\;
We report the results on ShapeNet~\cite{wu20153d} in Table \ref{table:shapenet}.
Similar to models constructed for experiments on ModelNet40, we experiment on a combination of different encoders and decoders.
As shown by the last column in Table \ref{table:shapenet}, the proposed hyperspherical module improves point cloud reconstruction consistently in all cases. 
When tasked on part segmentation, the point cloud decoders are removed from the models, and the embeddings concatenated with lifted point-wise features are processed by fully connected layers to predict part category. 
From the second column in Table \ref{table:shapenet}, part segmentation performance is not affected by the proposed module.
Multi-task learning results on shape reconstruction and part segmentation are shown in Figure \ref{fig:shapenet_multitask}.
By comparing the results of models with unconstrained embeddings, the proposed hyperspherical module have little effect on the performance of semantic tasks.
However, models with the proposed module have more stable performance when using large learning rates than those with unconstrained embeddings, since the same setting tend to cause the training with unconstrained embeddings unconverged.
In terms of converged results, models with our method still outperform their baselines with noticeable improvement.

\begin{figure*}[t!]
    \centering
    \includegraphics[width=0.9\textwidth]{{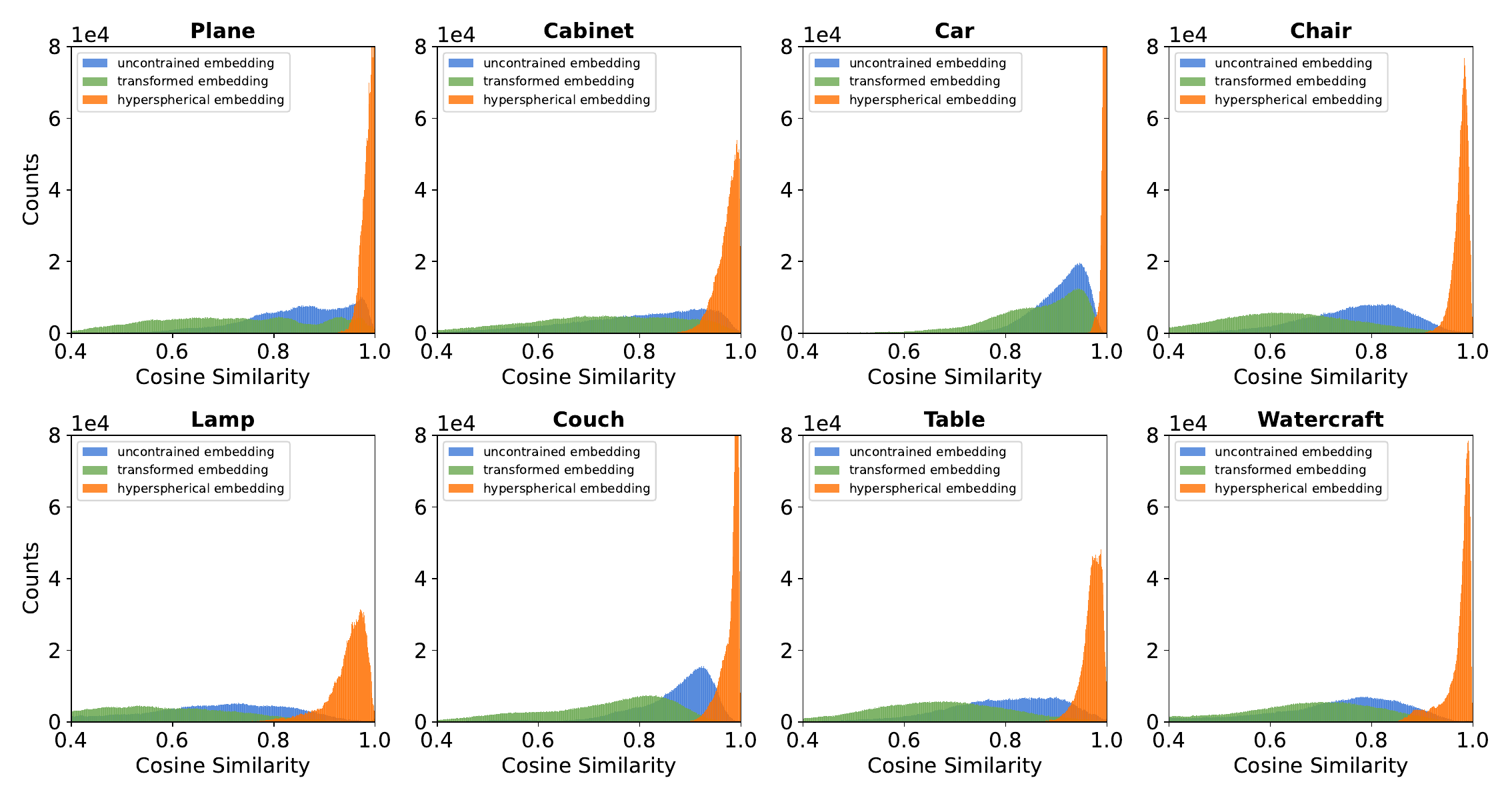}}
    \caption{Cosine similarity distribution of embeddings. We compute pairwise cosine distance between embeddings obtained from the test set in MVP dataset. We visualize the distribution of different classes as described in the plot titles. Hyperspherical embeddings have more compact angular distribution.}
    \label{fig:cosine_angular_distribution_all_class}
\end{figure*}

\begin{figure*}[t!]
    \centering
    \includegraphics[width=0.9\textwidth]{{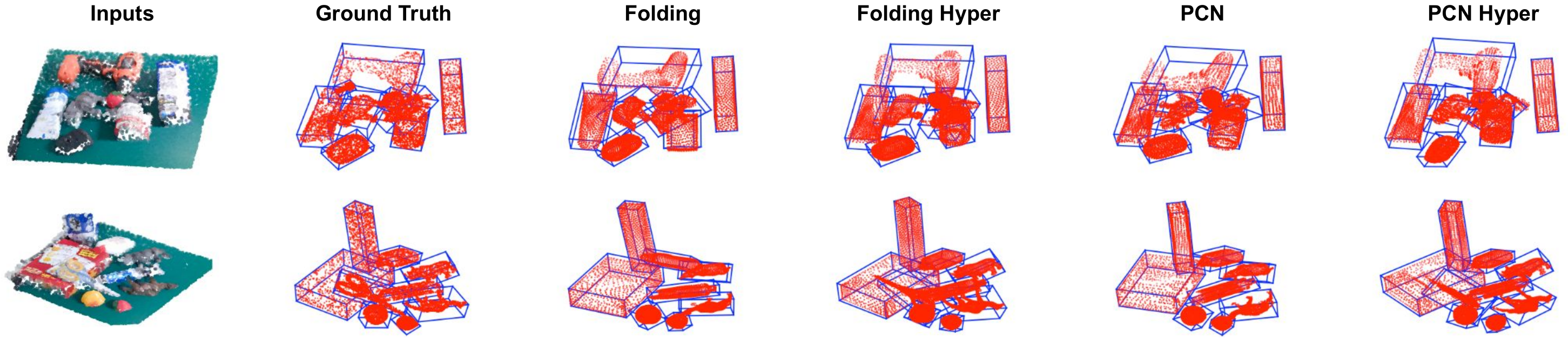}}
    \caption{More qualitative 3D detection, pose estimation, and point cloud completion results on GraspNet test set.}
    \label{fig:graspnet_more}
\end{figure*}

\subsection{More visualization}

We show the angular distribution of embeddings by computing the pairwise cosine similarity obtained from the test set in MVP dataset.
More visualizations of different classes as described in the plot titles are shown in Figure \ref{fig:cosine_angular_distribution_all_class}, and the distribution of overall classes is shown in Figure \ref{fig:cosine_similarity_gradients_between_tasks}

More qualitative results of 3D object detection, pose estimation, and point cloud completion can be found in Figure \ref{fig:graspnet_more}.

\end{document}